\documentclass[10pt,twocolumn,letterpaper]{article}

\usepackage[pagenumbers]{cvpr} 

\usepackage[dvipsnames]{xcolor}

\definecolor{cvprblue}{rgb}{0.21,0.49,0.74}
\usepackage[pagebackref,breaklinks,colorlinks,citecolor=cvprblue]{hyperref}

\usepackage{amssymb}
\usepackage{amsmath} 
\usepackage{graphicx} 
\usepackage[dvipsnames]{xcolor}
\usepackage{svg}
\usepackage{dblfloatfix}
\usepackage{float}

\def\paperID{*****} 
\def\confName{CVPR}
\def\confYear{2024}

\title{Gazebo Plants: Simulating Plant-Robot Interaction with Cosserat Rods}

\author{Junchen Deng\\
KAUST\\
{\tt\small junchen.deng@kaust.edu.sa}
\and
Samhita Marri\\
UIUC\\
{\tt\small marri2@illinois.edu}
\and
Jonathan Klein\\
KAUST\\
{\tt\small jonathan.klein@kaust.edu.sa}
\and
\phantom{---}Wojtek Pałubicki\\
\phantom{---}AMU\\
\phantom{---}{\tt\small wp06@amu.edu.pl}
\and
\phantom{--}Sören Pirk\\
\phantom{--}CAU\\
\phantom{--}{\tt\small sp@informatik.uni-kiel.de}
\and
Girish Chowdhary\\
UIUC\\
{\tt\small girishc@illinois.edu}
\and
Dominik L. Michels\\
KAUST\\
{\tt\small dominik.michels@kaust.edu.sa}
}

\begin{document}

\maketitle

\begin{abstract}\noindent
Robotic harvesting has the potential to positively impact agricultural productivity, reduce costs, improve food quality, enhance sustainability, and to address labor shortage.
In the rapidly advancing field of agricultural robotics, the necessity of training robots in a virtual environment has become essential.
Generating training data to automatize the underlying computer vision tasks such as image segmentation, object detection and classification, also heavily relies on such virtual environments as synthetic data is often required to overcome the shortage and lack of variety of real data sets.
However, physics engines commonly employed within the robotics community, such as ODE, Simbody, Bullet, and DART, primarily support motion and collision interaction of rigid bodies.
This inherent limitation hinders experimentation and progress in handling non-rigid objects such as plants and crops.
In this contribution, we present a plugin for the Gazebo simulation platform based on Cosserat rods to model plant motion.
It enables the simulation of plants and their interaction with the environment.
We demonstrate that, using our plugin, users can conduct harvesting simulations in Gazebo by simulating a robotic arm picking fruits and achieve results comparable to real-world experiments.

\noindent
\textbf{Supplemental Material:} An accompanying video can be found on \url{https://youtu.be/r8Q31w4bNbs} showing the experiments presented in this paper.


\noindent
\textbf{Plugin and Source Code:} Upon request, we are happy to share our \textsc{GazeboPlants} plugin open-source (MPL 2.0).



\end{abstract}

\section{Introduction}
Robotics has demonstrated its efficacy across a broad spectrum of applications, ranging from the assembly of automobiles in factories using manipulator arms like those manufactured by KUKA, to streamlining household chores such as vacuuming through the utilization of mobile robots like iRobot’s Roomba. The structured nature of these artificially created environments has propelled significant advancements in research within these domains over the past few decades. However, the progress of robotics in agricultural applications has been comparatively less conspicuous, primarily due to the inherent unstructured and cluttered characteristics of agricultural environments.
Nevertheless, there exists a pressing necessity for the advancement of agricultural robots to ensure food security for a growing global population in a sustainable manner \cite{foley2011solutions}, as well as to tackle issues like labor shortages.

While it is feasible to benchmark and assess different methods for a specific task within structured environments, such as picking a mug off a table, achieving a fair evaluation in agricultural settings is more challenging. Using the harvesting scenario as an illustration, several complexities arise. Firstly, no two plants, even of the same breed, possess identical structures. Secondly, once a fruit is harvested, reassessment becomes impractical. Thirdly, accessibility issues, such as cost or geography, can vary significantly among different plant varieties.

We believe that simulating plants will play an important role in addressing this disparity, akin to how progress in applications within structured environments was complemented by the development and accessibility of various simulation environments, as exemplified by \cite{brockman2016openai, xiazamirhe2018gibsonenv}.
Simulation tools for robotics contribute significant value, not only in the realm of robot learning but also in the crucial phase of verification before the deployment on physical robotic systems. This verification process is vital to ensuring the safety and functionality of both the robot and its operational environment. Particularly in data-intensive methods such as reinforcement learning, direct data collection on the robot may prove inefficient and potentially unsafe. Moreover, it encounters limitations in scalability, given the costs associated with employing multiple physical robots simultaneously to expedite the learning process. In contrast, simulation, as demonstrated in recent work like Issac Gym \cite{makoviychuk2021isaac}, allows several robots to learn in parallel and from each other, minimizing wear and tear on the robot and reducing risks to the environment. These simulation tools provide realistic sensing and testing capabilities that are not yet fully realized in agricultural settings.

The underlying computer vision tasks of agricultural robotics comprise, among others, image segmentation, object detection and classification. Addressing these tasks using state-of-the-art machine learning relies on the availability of sufficient training data. Such data can be generated in virtual environments resulting in large sets of synthetic data~\cite{klein2023synthetic}. The use of synthetic training data in agricultural robotics has already been successfully demonstrated, for example, for the particular application of harvesting sweet pepper~\cite{SweetPepper}.

From a more general perspective, according to Gartner\footnote{\url{https://techmonitor.ai/technology/ai-and-automation/ai-synthetic-data-edge-computing-gartner}}, the majority of training data will anyhow be synthetic by the end of 2024. In particular, for automatizing computer vision tasks, the quality of these synthetic data sets is crucial and requires accurate simulations in virtual environments.

In this paper, we introduce a novel simulation framework for plants integrated as a plugin within the Gazebo system \cite{koenig2004design}. Our simulation framework not only encompasses simulated plants that dynamically respond to robot interactions but also incorporates the crucial functionality of the robot manipulator in detaching fruits. Furthermore, we conduct comprehensive calibration tests across various parameters to ensure a realistic and accurate simulation environment.

\begin{figure}
  \centering
  \includegraphics[width=\linewidth]{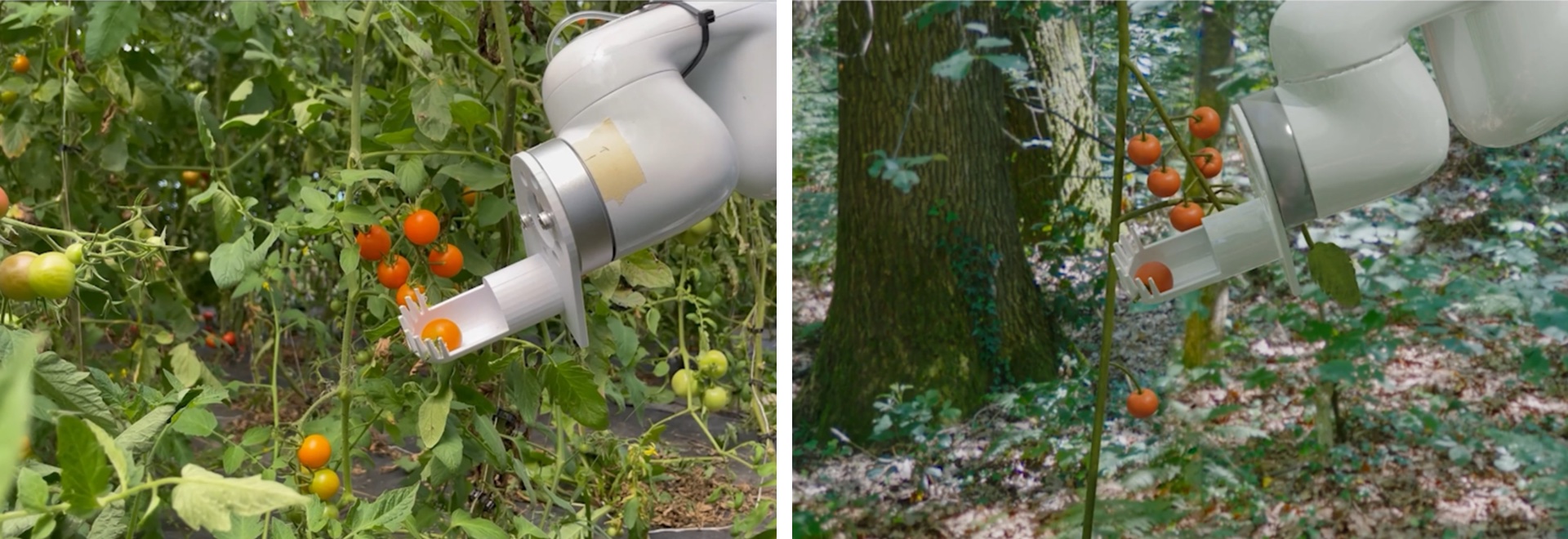}
  \caption{Visual comparison of simulation and real-world scenario: UFactory Lite 6 robotic arm picking cherry tomato using our custom gripper (left) and our corresponding simulation in the virtual environment (right).}
  \label{fig:gripper}
\end{figure}

\section{Related Work}
The relevant prior work includes contributions to plant simulation, virtual environments for robots, and the application of robots in agriculture.

\textbf{Plant Simulation.} Numerical methods for the simulation of plant dynamics can be divided into two categories based on the discretization approach.
The first one involves discretizing the plant into a mesh and using, for example, the Finite Element Method (FEM) \cite{10.1145/2461912.2461961,10.1145/2534329.2534357}, while the second one involves discretizing it into a skeleton and, among others, using Position-Based Dynamics (PBD) \cite{https://doi.org/10.1111/cgf.13326}.

To implement constraints between particles, PBD is often combined with Cosserat rods.
This classical model has been introduced first in visual computing by \cite{https://doi.org/10.1111/1467-8659.00594} to simulate more physically accurate bending and twisting effects. The classical model as well as several modern adaptations have been widely utilized for the simulation of fiber-like objects such as threads and hair \cite{https://doi.org/10.1111/1467-8659.00594,10.2312:SCA:SCA07:063-072,MICHELS2015136,10.1145/3072959.3073706,bertails:inria-00384718,10.1007/978-3-319-66320-3_22,10.1145/2508462,Shao:2021:GraphLearning,shao2022fluids,spillmann2008cosserat}.

In addition to the dynamics of the plant itself, some research focuses on other natural phenomena related to plants. This includes aspects such as plant growth 
 \cite{10.1145/2366145.2366188}, the effect of water in plants on plant morphology \cite{10.1111:cgf.12009,10.1145/3610548.3618218}, the formation process of complex root architectures \cite{Rhizomorph}, the combustion of plants \cite{10.1145/3130800.3130814}, the interaction between climate and vegetation \cite{10.1145/3528223.3530146}, the influence of the environment on plant shape \cite{10.1145/2185520.2185546,https://doi.org/10.1111/j.1467-8659.2009.01391.x,7836345,https://doi.org/10.1111/cgf.12736,https://doi.org/10.1111/cgf.13106}, the simulation of the cambium of trees \cite{10.1111:cgf.12566}, and modeling and simulating plant flowers \cite{10.1145/2897824.2925982,10.1145/3478513.3480548}. A botanical material model is proposed in \cite{10.1145/3072959.3073655} aiming for generating authentic simulations rooted in the biomechanics of trees. The development of complex ecosystems is simulated in \cite{SyntheticSilviculture}.

\textbf{Virtual Environments for Robots.} There are various simulation platforms such as Webots \cite{michel2004webotstm} and MuJoCo \cite{todorov2012mujoco} that are widely used in robotics.
We chose Gazebo \cite{koenig2004design} to integrate our plugin as it comes with easy integration with ROS, one of the standard frameworks for deploying robotic systems.
With additional support in ROS for various sensors such as vision and LIDAR, research in complex environments is made feasible.
There has also been enormous development to support different plugins in Gazebo such as articulated rigid objects interacting with fluid \cite{angelidis2022gazebo}, simulating radio frequency identification for industrial applications \cite{Alajami2022ros}, and transceiver for rescue operations \cite{ARVA}, which demonstrates its broad range of applications.  

\textbf{Robotic Applications in Agriculture.} Previous contributions tackle various agricultural tasks using different robots like harvesting \cite{shamshiri2018robotic}, phenotyping \cite{iqbal2020simulation}, and navigating in crop rows \cite{barbosa2021vision} by utilized simulation tools.
In \cite{shamshiri2018robotic}, sweet pepper plants are modeled in a virtual robot experimentation platform (VREP) to develop vision-based control methods in reaching a target pepper. However, the detachment of the pepper is not simulated to perform the harvesting task. In the case of \cite{iqbal2020simulation}, plant volume and height are estimated in ROS-Gazebo where the CAD models of plants are developed in SketchUp.
Using such synthetic data is more feasible than manual collection in the field where human errors can affect the ground truth.
Autonomous navigation in crop rows is also studied in \cite{iqbal2020simulation,barbosa2021vision} but both works are limited by the static nature of the CAD plant models as the dynamic behavior of plants, when external objects interact, is not simulated, causing a sim-to-real gap.

 
Other works include deploying artificial plant billboards like in \cite{taqi2017cherry}, which capture the visual appearance of the plant well, but fail to mimic its dynamic motion.
Another recent work \cite{Velasquez2022Predicting} developed a physical proxy for grasping apples but it does not account for interactions with other parts of the plant.
A more recent work \cite{barthelme2023robotic} utilized Unity's framework for the dynamic behavior of plants allowing robot-plant interaction in studying autonomous reaching of pruning targets, but the detachment of plant parts is not considered.
Our proposed work not only allows for robot-plant interactions but also simulates the detachment of fruits, further motivating researchers to develop autonomous harvesting techniques.

\textbf{Paper Outline.} The remaining sections of the paper are structured as follows: Section \ref{section:framework} covers the methodology behind our plugin as well as implementation details such as the interaction between the plugin and the simulation platform Gazebo.
Section \ref{section:experiments} comprises experiments, encompassing the impact of plant parameters on plant behavior, experiments related to fruit harvesting, and a comparison between simulation results and real-world outcomes. Concluding remarks are provided in Section \ref{section:conclusion}.

\begin{figure}[]
  \includegraphics[width=\linewidth]{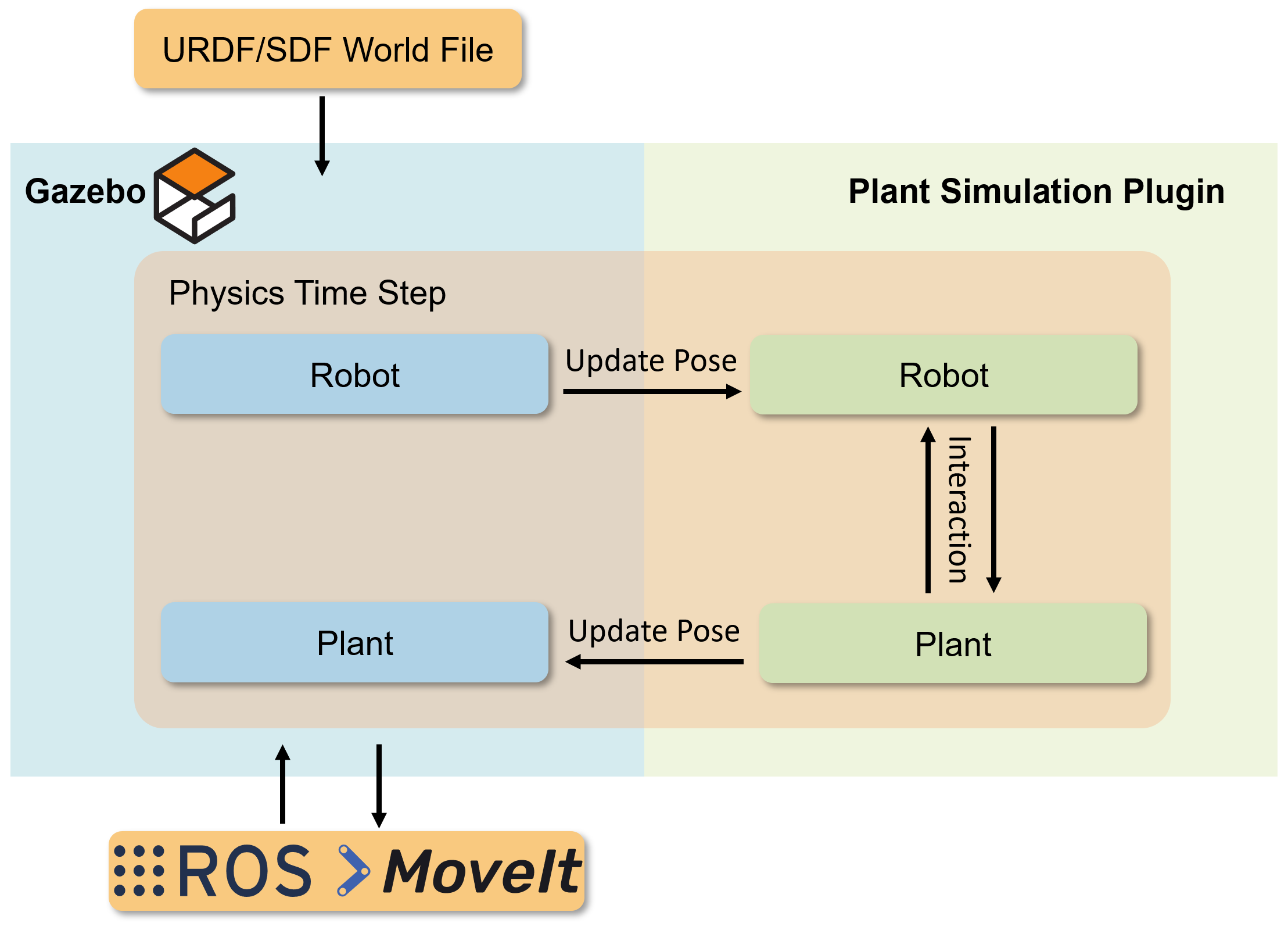}
  \caption{During initialization, Gazebo loads the world file, and within the parameters of our plugin, users need to specify both the plant file and the name of the robot intended to interact with the plant. Subsequently, the plugin proceeds to generate the plant as a combination of cylinders and spheres within the Gazebo environment, retrieving the robot's geometric data from Gazebo. During each physics time step, the following sequence of actions unfolds: Firstly, the robot's pose is updated, then it interaction with the plant is computed, and lastly, the pose of each individual component of the plant is reported back accordingly. Users can manipulate the robot by interfacing with Gazebo through widely used libraries such as ROS and MoveIt.}
  \label{fig:pipeline}
\end{figure}


\section{Methodology}\label{section:framework}

\subsection{Interaction Framework}

We developed our plant simulation plugin by utilizing the Gazebo World plugin as its foundation.
Within Gazebo's world file, users are required to specify the robot's name, which will interact with the plant.
In the initialization phase, the plugin then retrieves the geometric information of the robot object from Gazebo and subsequently constructs data structures for collision detection.
It proceeds to generate the plant model using a combination of cylinders and spheres.
The whole pipeline is shown in Figure~\ref{fig:pipeline}.
Given that we employ a position-based simulation method, the plugin only needs to periodically request the specific robot's pose information from Gazebo at each time step.
Since our implementation takes the form of a Gazebo plugin, users maintain the versatility to utilize our plugin with any library that possesses the capability to interface with Gazebo, such as ROS and MoveIt.

\subsection{Position and Orientation Based Cosserat Rods}

\begin{figure}[]
  \includegraphics[width=\linewidth]{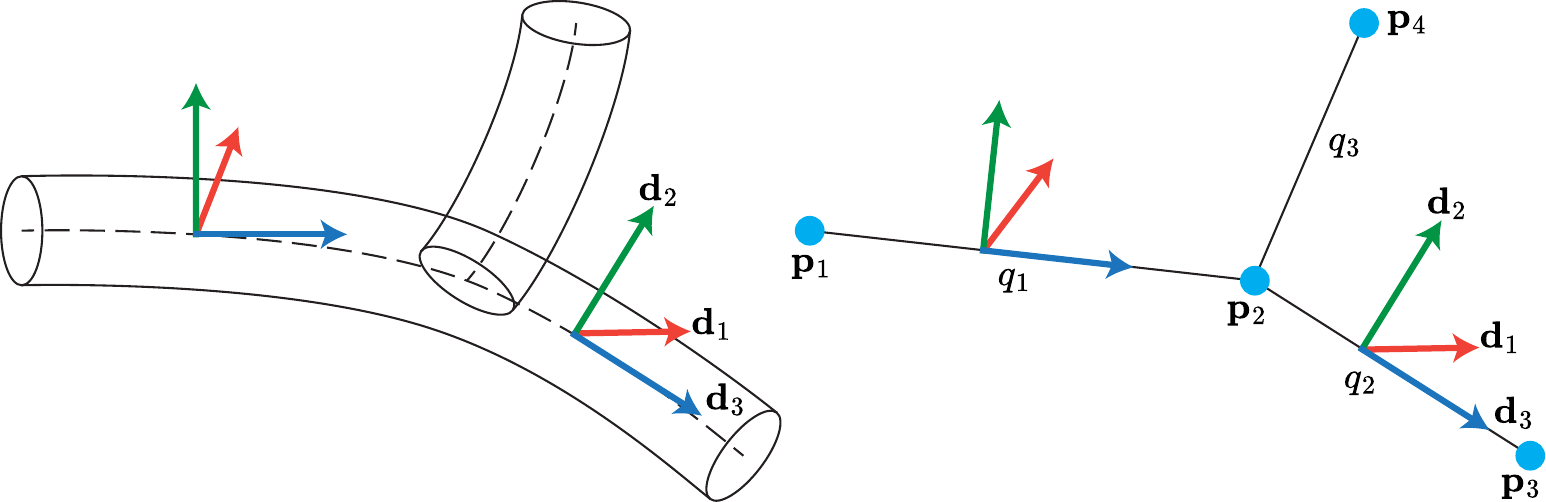}
  \caption{Illustration of the plan discretization in which $\{\mathbf{d}_{k}\}$ denotes the local coordinate system for the Cosserat rod.}
  \label{fig:discrete}
\end{figure}

We choose to adopt the Position-Based Dynamics (PBD) framework, as detailed in \cite{10.2312:sca.20161234} to model plant branches as Cosserat rods. The Cosserat theory models a rod as the curve $\mathbf{r}(s):\mathbb{R}\rightarrow\mathbb{R}^{3}$ along the centerline. As shown on the left in Figure~\ref{fig:discrete}, for each point on the curve, there is an orthonormal frame with basis $\{\mathbf{d}_{1}(s), \mathbf{d}_{2}(s), \mathbf{d}_{3}(s)\}$, where $\mathbf{d}_{3}(s)$ is the tangent direction of the curve. To establish this coordinate system, we rely on a quaternion $q(s)$, which serves as the means to rotate the basis vectors from the world coordinate system. The strain measure for bending and twisting is determined using the Darboux vector $\Omega(s)$ which is defined as 
\begin{align*}
\Omega(s) = 2\bar{q}(s)\frac{\partial}{\partial s}q(s)\,,
\end{align*}
where $\bar{q}$ denotes the conjugate quaternion. Because some rods have an initial bending and torsion, we denote the rest pose Darboux vector as $\Omega(s)^{0}$. Thus, the strain measure of bending and torsion is defined as 
\begin{align*}
\Delta\Omega(s) = \Omega(s) - \Omega(s)^{0}\,.
\end{align*}

To discretize the curves, each rod element comprises two adjacent particles and a quaternion that holds both the position and orientation information of the coordinate system, as shown in Figure~\ref{fig:discrete}.
The Darboux vector should be also discretized between each neighbor rod. The derivative of the quaternion is discretized by a finite difference approximation ${\partial q_{i}}/{\partial s} =  (q_{i+1} - q_{i})/{l_{i}}$, where $l_{i}$ denotes the average length of neighbor rods $i$ and $i+1$.
Thus, the discrete Darboux vector is defined as
\begin{align*}
\Omega_{i}=\frac{2}{l_{i}}(\bar{q_{i}}q_{i+1})\,.
\end{align*}
Here, to handle the ambiguity introduced by dropping the real part of the quaternion in 
\cite{10.2312:sca.20161234}, we employ the augmented Darboux vector \cite{Hsu2023} as above. 

Within the PBD framework, we adopt a pair of constraints to enforce the desired behavior of the plant.
Specifically, we make use of a stretch and shear constraint $\mathbf{C}_{s}$.
This constraint is designed to approximately preserve the rod's length and relies on the configuration of two endpoints $\mathbf{p}_{1}$ and $\mathbf{p}_{2}$, and on an its length and original orientation $\mathbf{d}_{3}$:
\begin{align*}
\mathbf{C}_{s}(\mathbf{p}_{1}, \mathbf{p}_{2}, \mathbf{d}_{3})=\frac{1}{l_{0}}(\mathbf{p}_{2}-\mathbf{p}_{1})-\mathbf{d}_{3}\,,
\end{align*}
where $l_{0}$ represents the rest length of each rod.
To simulate bending and twisting behavior, we use constraint $\mathbf{C}_{b}$, which is defined based on the difference between the Darboux vector $\mathbf{\Omega}$ at the current pose and the rest pose:
\begin{align*}
\mathbf{C}_{b}(q_{1}, q_{2})=\mathbf{\Omega}-s\mathbf{\Omega}_{0}\,,
\end{align*}
\begin{align*}
s = 
\begin{cases}
+1 & \text{for }|\mathbf{\Omega}-\mathbf{\Omega}_{0}|^{2}<|\mathbf{\Omega}+\mathbf{\Omega}_{0}|^{2}\,,\\
-1 & \text{for }|\mathbf{\Omega}-\mathbf{\Omega}_{0}|^{2}>|\mathbf{\Omega}+\mathbf{\Omega}_{0}|^{2}\,.
\end{cases}
\end{align*}
This adjustment of $s$ is necessary to move the rod to the nearest rest pose since $\mathbf{\Omega}$ and $-\mathbf{\Omega}$ represent the same rotation.

\subsection{Collision Detection and Response}
In our simulation, we represent the rod as a capsule body, with radius information stored at each node.
To simplify the modeling of fruits, we model them as spheres that are attached to a specific node. 

\begin{figure}[]
  \includegraphics[width=\linewidth]{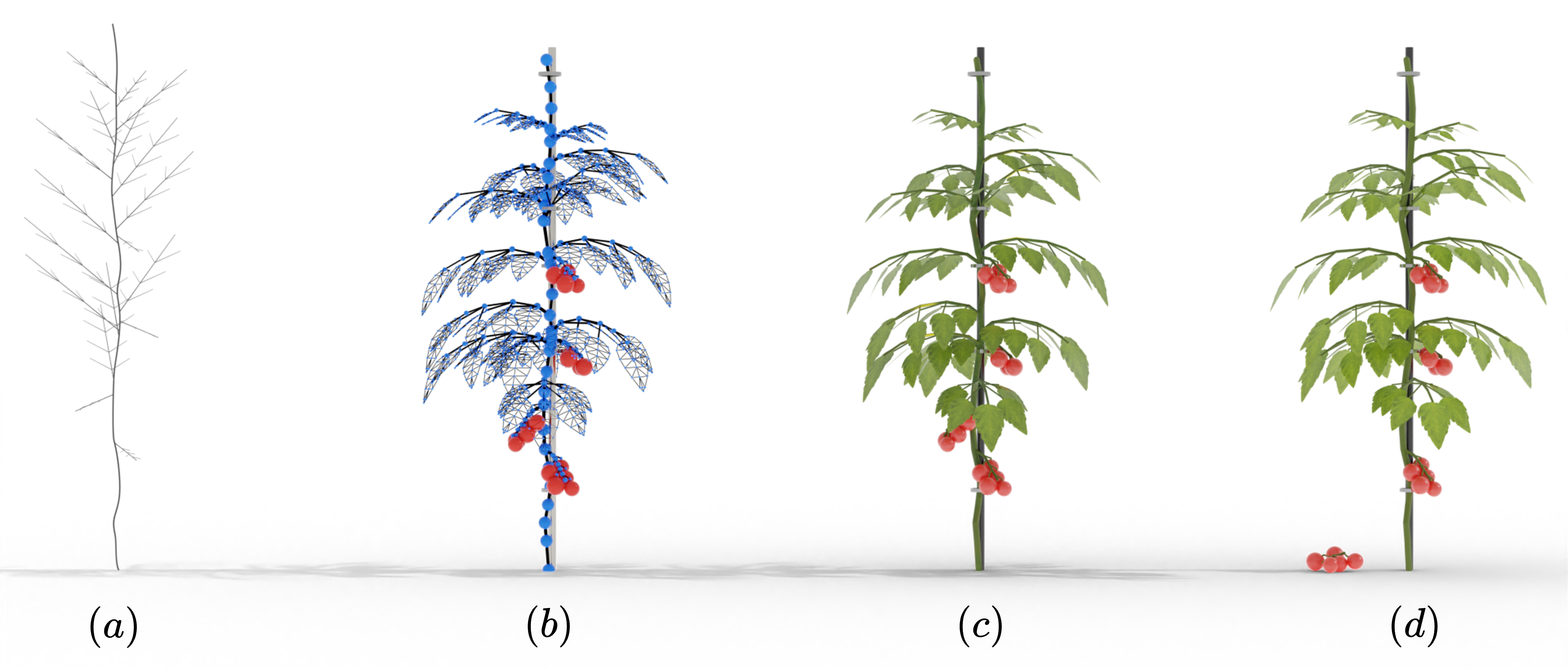}
  \caption{(a) portrays a plant through the use of multiple curves. Through the process of sampling along these curves, connecting the branches, and introducing leaves, we can transform it into (b), which represents a plant through a collection of particles linked by a graph. (c) depicts the plant, as represented by (b), in the physical world. (d) illustrates the action of harvesting a cluster of fruits from the plant.}
  \label{fig:overview}
\end{figure}

Resolving collisions involves separating the two colliding objects by pushing them away from each other along the normal vector at the point of collision. When dealing with self-collisions, the collision pairs include capsule-capsule, capsule-sphere, and sphere-sphere interactions. We calculate the collision position and its corresponding normal following the methodology outlined in \cite{10.5555/1121584}. 

We implement the robot/plant interaction as a one-way coupling, where the robot can move the plant but the plant does not affect the robots movement.
This is a reasonable assumption for crop plants but may be less suitable for timber scenarios.
The collision pairs in the crop context then include interactions between the sphere and the robotic arm, as well as between the capsule and the robotic arm. To handle collisions with the robotic arm, we utilize a Signed Distance Field (SDF)\cite{Discregrid_Library}.
This approach enables us to efficiently determine the closest distance from any point to the robotic arm and obtain the corresponding normal direction. In the case of collisions involving the capsule body and the robotic arm, we performed equidistant sampling along the main axis of the capsule body.
Subsequently, we identified the location of the collision as the point with the closest distance among all the sampled points.

\begin{figure}
  \includegraphics[width=\linewidth]{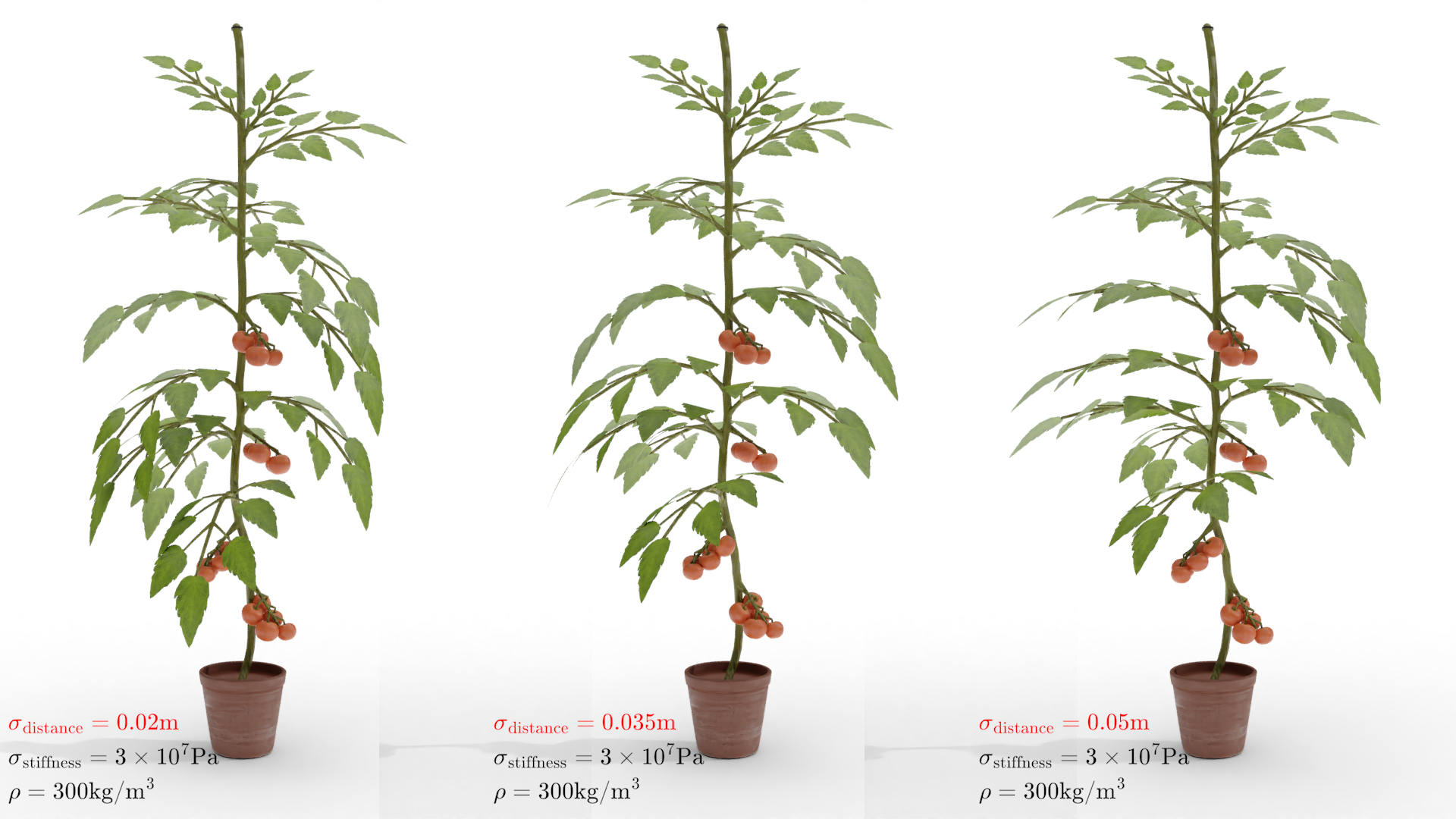}
  \caption{Plant behaviors with different $\sigma_{\text{distance}}=0.02, 0.035, 0.05~\text{m}$. For all three results, $\sigma_{\text{stiffness}}=2\times 10^{4}$ and $\rho=300~\text{kg}/\text{m}^{3}$.}
  \label{fig:resolution}
\end{figure}

\begin{figure}
  \includegraphics[width=\linewidth]{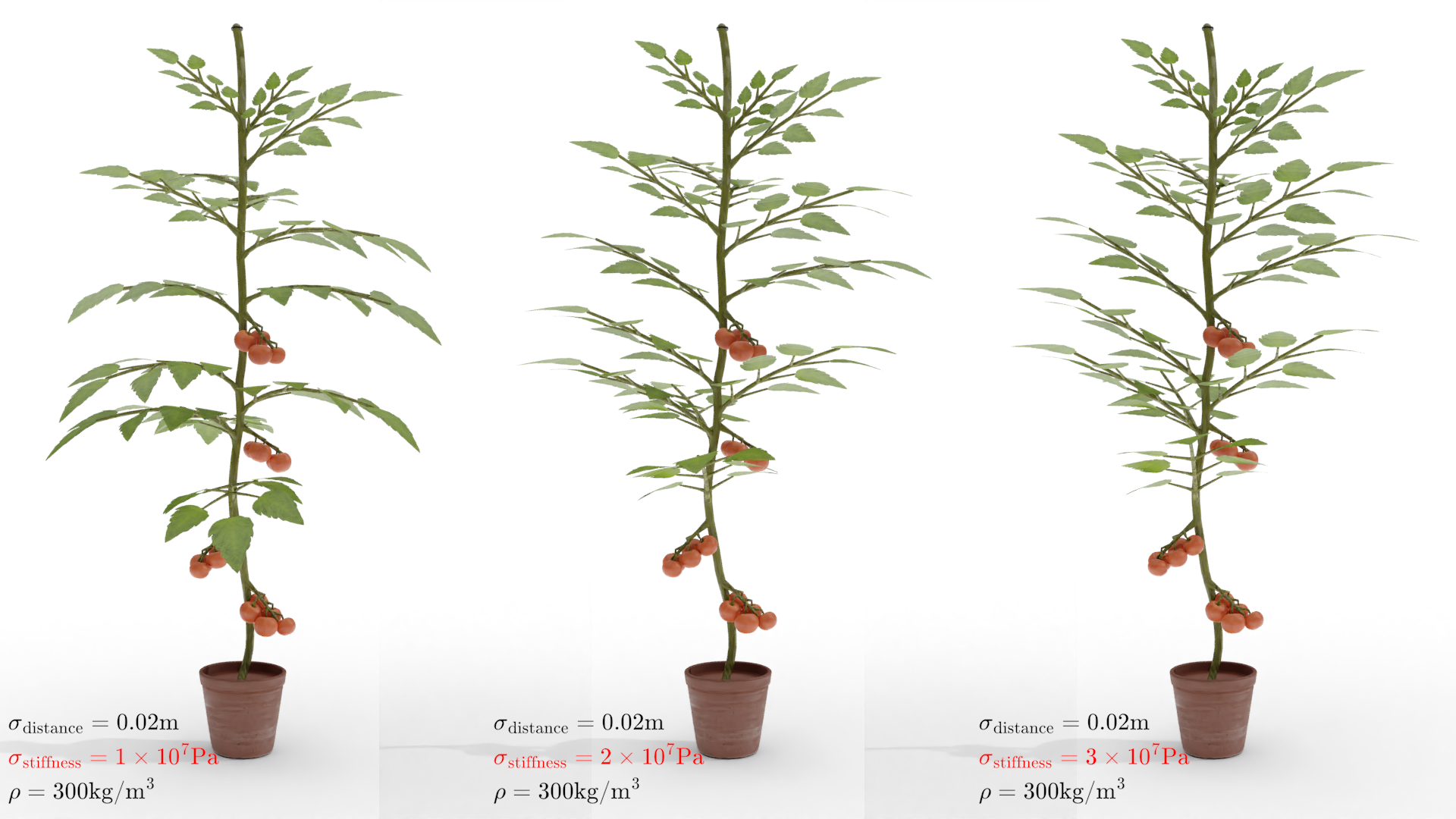}
  \caption{Plant behaviors with different $\sigma_{\text{stiffness}}=1, 2, 3\times 10^{7}~\text{Pa}$. For all three results, $\sigma_{\text{distance}}=0.02~\text{m}$ and $\rho=300~\text{kg}/\text{m}^{3}$.}
  \label{fig:stiffness}
\end{figure}

\subsection{Plant Fracture}

Our framework supports plant fracture (Figure~\ref{fig:overview}.d) through the assessment of strain.
When a stretch or bending constraint value $\mathbf{C}_{s}$ or $\mathbf{C}_{b}$ exceeds a predefined threshold $\mathbf{C}_{s}^{\max}$ or $\mathbf{C}_{b}^{\max}$, the corresponding rod element between the particles is fractured and removed from the system.
Furthermore, any constraints associated with this particular rod are also eliminated.

\subsection{Plant Model}

We model plants by hand drawing curves for individual branches in 3D space (Figure~\ref{fig:overview}.a).
We then densely sample these curves using a step size smaller than the closest distance between any two curves.
We establish a threshold, denoted as $\sigma_{\text{connect}}$, to add the branch connections.
Nodes on one branch are connected to the endpoints of the nearest other branches when their distance is less than this threshold. This connection process is performed through breadth-first search, commencing from the specified root node, ensuring that the resulting connected graph forms a tree structure.

\begin{figure}
  \includegraphics[width=\linewidth]{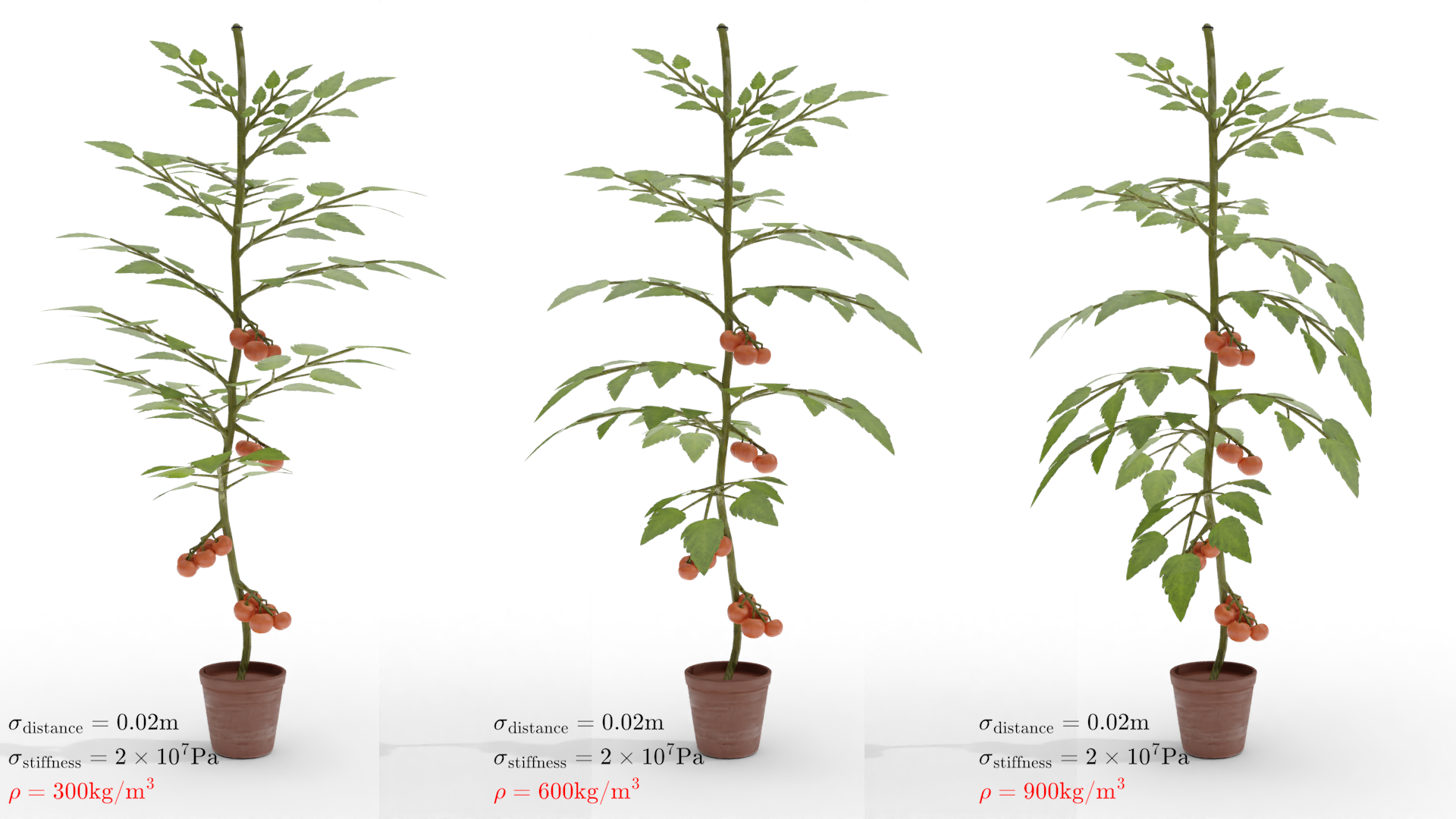}
  \caption{Plant behaviors with different $\rho=300, 600, 900~\text{kg}/\text{m}^{3}$. For all three results, $\sigma_{\text{distance}}=0.02~\text{m}$ and $\sigma_{\text{stiffness}}=2\times10^{7}~\text{Pa}$.}
  \label{fig:density}
\end{figure}

Next, we simplify the nodes within this tree by setting a distance threshold, denoted as $\sigma_{\text{distance}}$. Initially, certain nodes must be preserved, including junction nodes (nodes with a degree greater than 2) and leaf nodes (nodes with a degree equal to 1), referred to as key nodes. Subsequently, a depth-first search begins from the root node, and if the distance $d$ from a node $u$ to a preserved node $v$ in its ancestry is less than $\sigma_{\text{distance}}$, then node $u$ is also preserved. For key nodes, if their $d$ is less than $0.5\,\sigma_{\text{distance}}$ and $v$ is not a key node, then node $v$ is no longer retained. This approach prevents the issue of two retained nodes being too close to each other. Finally, we specify certain nodes for adding leaves or fruits, and we have predefined leaf templates for adding leaves.

\begin{figure*}[]
\centering  
  \includegraphics[width=1\textwidth]{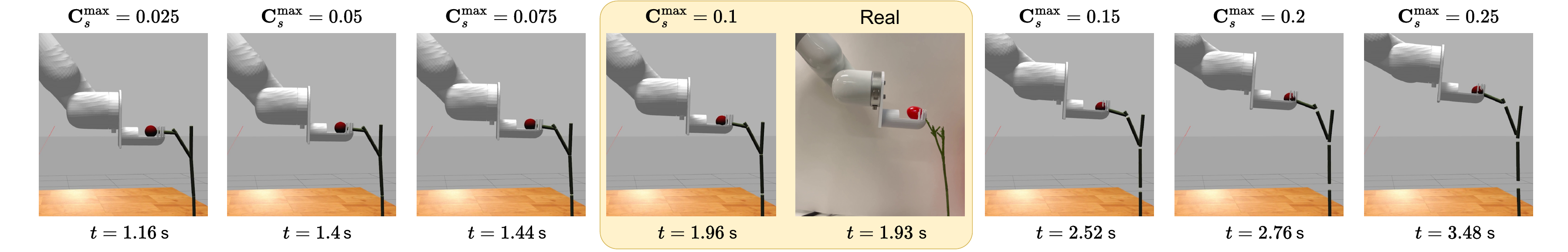}
  \caption{Performing a calibration experiment to identify the threshold for fruit detaching in Gazebo by comparing to a real demonstration where the robot moves upwards to detach the fruit. It is ensured that the manipulator's trajectories in the real demonstration and simulation are similar. The snapshots shown here are taken just before the moment of detachment for various stress thresholds $\mathbf{C}_{s}^{\max}$ (labeled at the top) with the corresponding time taken to detach (labeled at the bottom). We observe that the detachment time in simulation is closest to the real demonstration at $\mathbf{C}_{s}^{\max} = 0.1$, and as $\mathbf{C}_{s}^{\max}$ increases or decreases, the detachment time also increases or decreases respectively. Qualitatively, we also notice that the shape of the plant in simulation matches closest to the real demonstration at $\mathbf{C}_{s}^{\max} = 0.1$. It is particularly noticeable at a higher $\mathbf{C}_{s}^{\max}=0.25$, where the plant in simulation stretches further than at $\mathbf{C}_{s}^{\max} = 0.1$.}
  \label{fig:calibration_cmax_upward}
\end{figure*}

\begin{figure*}[]
\centering
  \includegraphics[width=0.985\textwidth]{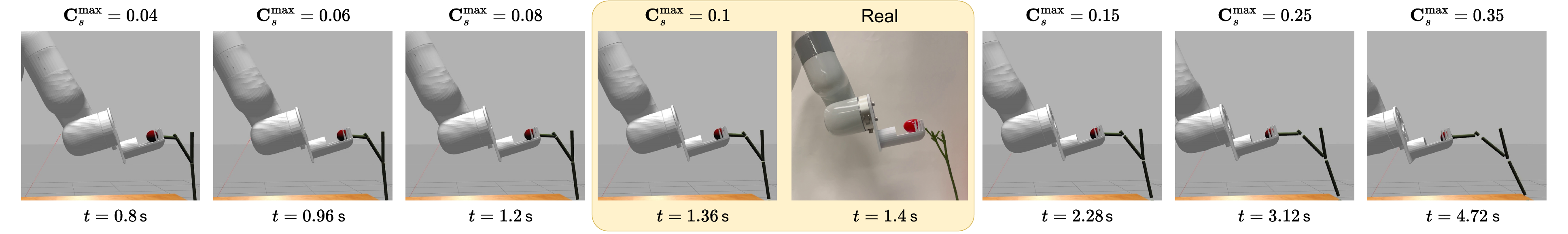}
  \caption{Performing a calibration experiment to identify the threshold for fruit detaching in Gazebo by comparing to a real demonstration where the robot moves horizontally away from the plant. It is ensured that the manipulator's trajectories in the real demonstration and simulation are similar. The snapshots shown here are taken just before the moment of detachment for various stress thresholds $\mathbf{C}_{s}^{\max}$ (labeled at the top) with the corresponding time taken to detach (labeled at the bottom). We observe that the detachment time in simulation is closest to the real demonstration at $\mathbf{C}_{s}^{\max} = 0.1$, and as $\mathbf{C}_{s}^{\max}$ increases or decreases, the detachment time also increases or decreases respectively. Qualitatively, we also notice that the shape of the plant in simulation matches closest to the real demonstration at $\mathbf{C}_{s}^{\max} = 0.1$. It is particularly noticeable at a higher $\mathbf{C}_{s}^{\max}=0.35$, where the plant in simulation stretches further than at $\mathbf{C}_{s}^{\max} = 0.1$.}
  \label{fig:calibration_cmax_horizontal}
\end{figure*}

Now, we represent the plant as a graph structure, where each edge in this graph corresponds to a rod, as illustrated in Figure~\ref{fig:overview}.b. Even when dealing with leaves, rather than for a representation using individual triangles, we also treat their mesh as a graph. Each rod is subject to both stretch and shear constraints, and for every pair of adjacent rods, we add bend and twist constraints.

\section{Experimental Results}\label{section:experiments}

\subsection{Plant Parameters}
We conduct several experiments to demonstrate that various plant parameters can influence the stiffness exhibited by plants. These parameters include $\sigma_{\text{distance}}$, used during the simplification of the plant model, the plant's stiffness parameter $\sigma_{\text{stiffness}}$, and the density of the plant's stem $\rho$.

In the Cosserat rod model, the stiffness of a plant is closely related to the number of nodes it possesses.
When trees have a greater number of nodes, the system's degrees of freedom increase, resulting in an overall softer behavior of the plant.
We have the ability to manage the number of nodes by modifying the distance parameter $\sigma_{\text{stiffness}}$ during the tree creation process.
As depicted in Figure~\ref{fig:resolution}, a smaller distance yields a greater number of nodes, causing the branches to exhibit more sag due to the influence of gravity.

In addition to the indirect influence of the number of nodes, we also have a direct parameter $\sigma_{\text{stiffness}}$ that allows us to set Young's modulus at each node of the plant. This direct parameter gives us control over the stiffness of the plant. For herbaceous plants, $\sigma_{\text{stiffness}}$ can be set in the order of $10^6\sim10^7\text{Pa}$, while for woody plants, it can be set in the order of $10^8\sim10^{10}\text{Pa}$. As depicted in Figure~\ref{fig:stiffness}, from left to right, an increase in $\sigma_{\text{stiffness}}$ results in the branches exhibiting a straighter behavior.

To align the plant with the target, the user will also need to specify the density of the plant's material.
In the case of herbaceous plants, the typical density falls within the range of $200$ to $1000~\text{kg}/\text{m}^{3}$. In Figure~\ref{fig:density}, we present the results with three distinct density values.
A larger plant density leads to an increased bending under the influence of gravity and to a reduced susceptibility to external forces such as wind.

All experiments are conducted on a workstation with NVIDIA RTX 3090 GPU and Intel Xeon Gold 6136 3.00~GHz CPU. For $\sigma_{\text{distance}}=0.02~\text{m}$, the maximum number of rods in the scene is 2\,242, with a 0.93 simulation to real-time ratio.

\begin{figure*}[]
\centering
  \rotatebox[x=0em, y=0.81em]{+90}{\small{Experiment\;\;\;\;\;\;Simulation}}\hfill{}\includegraphics[width=0.975\textwidth]{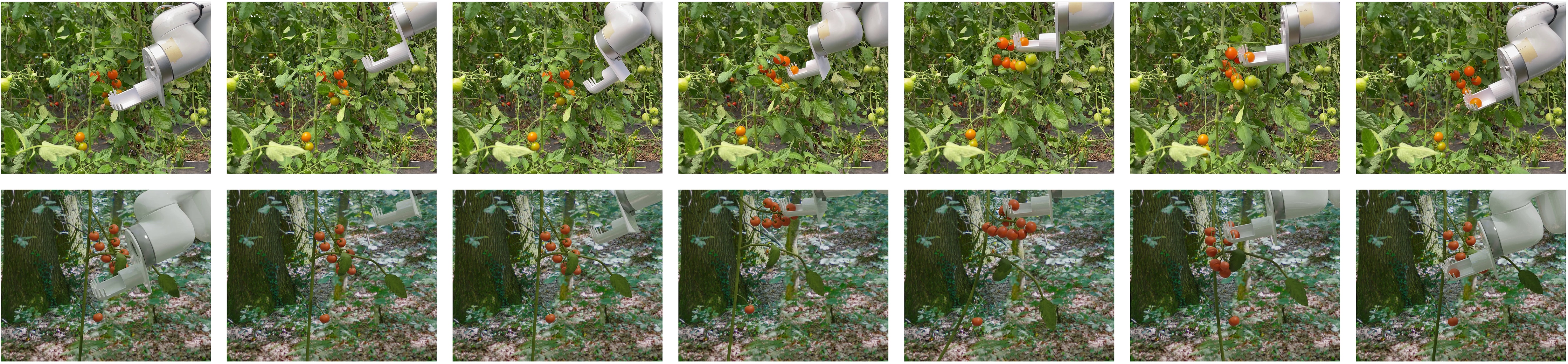}\\
  \includegraphics[width=1.0\textwidth]{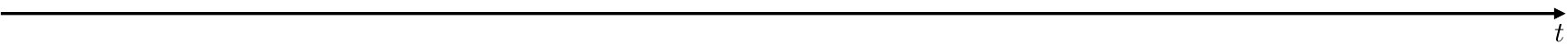}
  \caption{Selected frames extracted from the real-world harvesting sequence (top row) demonstrate the actual tomato retrieval process using a robotic arm in a greenhouse. The corresponding frames at the bottom row depict our simulation results, showing the similarity between simulated and real-world harvesting scenarios.}
  \label{fig:teaser-cmp}
\end{figure*}

\subsection{Calibration for Fruit Detachment}

Since fruit detachment depends on the value of the strain constraint threshold $\mathbf{C}_{s}^{\max}$, we perform a calibration using real world measurements.
In particular, we perform two calibration experiments: The first one involves detaching the fruit in an upward motion of the robot (Figure~\ref{fig:calibration_cmax_upward}), and the second one involves a horizontal motion away from the plant (Figure~\ref{fig:calibration_cmax_horizontal}).
For the demonstration of fruit detaching using the UFactory Lite 6 robotic arm, we used an artificial cherry tomato branch and a custom gripper as shown in Figure~\ref{fig:gripper}.
We chose this gripper design over traditional scissors-based such as \cite{jun2021towards} to avoid causing damage to the plant.
We measure the time to detach the fruit from the start of the robot's motion as reported at the bottom of each snapshot in Figure~\ref{fig:calibration_cmax_upward} and Figure~\ref{fig:calibration_cmax_horizontal}.

In both cases, we notice that at $\mathbf{C}_{s}^{\max}=0.1$, the time taken to detach on the real system and in the simulation are similar. As seen in the figures, as the value of the stress threshold increases, $\mathbf{C}_{s}^{\max} >0.1$, there is an increase in the stretch of the plant, and for lower thresholds, the stretch is not as pronounced.

\subsection{Comparison with Reality}
To demonstrate the practicality of our simulator, we conducted a comparison between real-world outcomes and simulated results. Employing a robotic arm, we harvested a tomato in a greenhouse. Subsequently, we constructed a model of a similar plant and, by adjusting its parameters, successfully aligned the simulation results with the real-world outcomes in Figure~\ref{fig:teaser-cmp}. 

As the tomato plant which we have harvested has been fixed from the upper end and suspended from a beam, attempting to pick the fruit by pulling would result in a significant displacement of the entire plant, exceeding the control range of the robotic arm. Consequently, we opted for a bending method to harvest the fruit.
During the simulation, we set the bending threshold between the two segments connected to the fruit $\mathbf{C}_{b}^{\max}$ to 0.31.
The full experiment is shown in the accompanying video.


\section{Conclusion}\label{section:conclusion}
In this project, we have developed a plant simulator plugin for Gazebo, based on Position-Based Dynamics (PBD) and Cosserat rods.
The plugin facilitates interaction between robots and plants, especially in the context of fruit harvesting.
We support two methods for fruit picking: Stretching and bending. Compared to existing harvesting robots that use scissors, the stretching method closely resembles human picking, causing less damage to plants.
On the other hand, the bending method allows for harvesting within smaller spaces. The comparison of simulation results with real-world outcomes validates the accuracy of our approach.
We believe that our plugin enables users to train their harvesting robots, not limited to cherry tomatoes, for autonomous picking.
This has the potential to enhance productivity and reduce labor costs by allowing robots to autonomously harvest a variety of crops.

While our plugin is a step towards realistic simulation of plants, there are limitations:
The plant sags due to gravity instead of staying in the position we modeled. To stay in the initial position requires a large hardness value, which affects the dynamics of the plant; a better approach would be to model the plant to fall due to gravity, but this effect is not linear and requires the user to change the initial position of the plant several times.
To solve this problem, the initial Darboux vector should be tuned by optimization.

\section*{Acknowledgements}
This work has been partially supported by KAUST through the baseline funding of the Computational Sciences Group. 




\bibliographystyle{IEEEtran}
\bibliography{main}

\end{document}